\pdfoutput=1

\documentclass[11pt]{article}

\usepackage[]{EMNLP2023}

\usepackage{times}
\usepackage{latexsym}

\usepackage[T1]{fontenc}

\usepackage[utf8]{inputenc}
\usepackage{cjhebrew}

\usepackage{microtype}

\usepackage{inconsolata}

\usepackage{multirow}
\usepackage{tabularx}

\usepackage{graphicx}
\usepackage{subfigure}

\title{Explicit Morphological Knowledge Improves \\ Pre-training of Language Models for Hebrew}

\author{
    Eylon Gueta \hfill Omer Goldman \hfill Reut Tsarfaty\\
    Computer Science Department, Bar Ilan University \\
    \texttt{\{guetaeylon, omer.goldman, reut.tsarfaty\}@gmail.com}
}

\begin{document}
\maketitle
\begin{abstract}
Pre-trained language models (PLMs) have shown remarkable successes in acquiring a wide range of linguistic knowledge, relying solely on self-supervised training on text streams.
Nevertheless, the effectiveness of this language-agnostic approach has been frequently questioned for its sub-optimal performance when applied to morphologically-rich languages (MRLs).
We investigate the hypothesis that incorporating explicit morphological knowledge in the pre-training phase can improve the performance of PLMs for MRLs.
We propose various morphologically driven tokenization methods enabling the model to leverage morphological cues beyond raw text. 
We pre-train multiple language models utilizing the different methods and evaluate them on Hebrew, a language with complex and highly ambiguous morphology.
Our experiments show that morphologically driven tokenization demonstrates improved results compared to a standard language-agnostic tokenization, on a benchmark of both semantic and morphologic tasks.
These findings suggest that incorporating morphological knowledge holds the potential for further improving PLMs for morphologically rich languages.
\end{abstract}

\section{Introduction}
Pre-trained language models (PLMs) have achieved state-of-the-art results for a great variety of tasks by utilizing a language-agnostic approach of learning representations from raw text \cite{devlin-etal-2019-bert, radford2018improving}.
This general approach enables the pre-training of PLMs in languages of different characteristics, either as monolingual \cite{seker-etal-2022-alephbert, antoun-etal-2020-arabert} or multilingual PLMs \cite{conneau-etal-2020-unsupervised, xue-etal-2021-mt5}.
Most PLMs rely on a statistically driven tokenization process, e.g. WordPiece or BPE \cite{schuster2012japanese, sennrich-etal-2016-neural}, which is responsible for converting raw-text into a finite set of symbols --- the fundamental units of PLM training.

Despite being highly effective for many languages, the Hebrew language, as an example of  morphologically-rich language (MRL) with a highly ambiguous and fusional morphology, introduces challenges to these prevailing tokenization methods, which makes them sub-optimal for MRLs \cite{tsarfaty-etal-2019-whats, tsarfaty-etal-2020-spmrl, cao-rimell-2021-evaluate, mager-etal-2022-bpe, araabi-etal-2022-effective}.

\begin{table*}[t]
\resizebox{\textwidth}{!}{%
\begin{tabular}{ccccccc}
\multicolumn{1}{c|}{\textbf{Prefix}} &
  \multicolumn{1}{c|}{\textbf{Suffix}} &
  \multicolumn{1}{c|}{\textbf{Form}} &
  \multicolumn{1}{c|}{\textbf{\begin{tabular}[c]{@{}c@{}}English\\ Translation\end{tabular}}} &
  \multicolumn{1}{c|}{\textbf{\begin{tabular}[c]{@{}c@{}}WordPiece\\ Tokenization\end{tabular}}} &
  \multicolumn{1}{c|}{\textbf{\begin{tabular}[c]{@{}c@{}}Morphological\\ Segmentation\\ Tokenization\end{tabular}}} &
  \textbf{\begin{tabular}[c]{@{}c@{}}Prefix-Suffix\\ Separation\\Tokenization\end{tabular}} \\ \hline
\multicolumn{1}{c|}{-} &
  \multicolumn{1}{c|}{-} &
  \multicolumn{1}{c|}{\</s.hrwr>} &
  \multicolumn{1}{c|}{liberation} &
  \multicolumn{1}{c|}{\</s.hrwr>} &
  \multicolumn{1}{c|}{\underline{\</s.hrwr>}} &
  \underline{\<.hrwr>}+\</s> \\
\multicolumn{1}{c|}{\</s>} &
  \multicolumn{1}{c|}{-} &
  \multicolumn{1}{c|}{\</s/s.hrwr>} &
  \multicolumn{1}{c|}{that lib.} &
  \multicolumn{1}{c|}{\<wr>\#\# \</s/s.hr>} &
  \multicolumn{1}{c|}{\underline{\</s.hrwr>}+\</s>} &
  \</s.hrwr>+\</s> \\
\multicolumn{1}{c|}{\<w>} &
  \multicolumn{1}{c|}{\<h>} &
  \multicolumn{1}{c|}{\<w/s.hrwrh>} &
  \multicolumn{1}{c|}{and her lib.} &
  \multicolumn{1}{c|}{\<h>\#\# \<w/s.hrwr>} &
  \multicolumn{1}{c|}{\<h>+\underline{\</s.hrwr>}+\<w>} &
  \<h>+\underline{\<.hrwr>}+\</s>+\<w> \\
\multicolumn{1}{c|}{\<k200C>+\<w>} &
  \multicolumn{1}{c|}{\<nw>} &
  \multicolumn{1}{c|}{\<wk/s.hrwrnw>} &
  \multicolumn{1}{c|}{and as our lib.} &
  \multicolumn{1}{c|}{\<nw>\#\# \<.hrwr>\#\# \<wk/s>} &
  \multicolumn{1}{c|}{\<nw>+\underline{\</s.hrwr>}+\<k200C>+\<w>} &
  \<w>+\<n200C>+\underline{\<.hrwr>}+\</s>+\<k200C>+\<w> \\
(1) &
  (2) &
  (3) &
  (4) &
  (5) &
  (6) &
  (7)
\end{tabular}%
}
\caption{Tokenization methods: the word \</s.hrwr> (liberation, abbreviated as \textit{lib.}) in its original form and when joined with a mixture of prefixes and suffixes.
For each tokenization method, the host's overlapping subword is underlined, highlighting differences of our examined tokenization methods.
Tokenization is based on \cite{guetta2022large} which has the extremely large vocabulary size of 128K subwords, the biggest upon all existing Hebrew PLMs.}
\label{table:tok_methods}
\end{table*}

In MRLs, linguistic information is reflected in the modification of word forms rather than added functional word, as is the case in configurational languages such as English.
Table \ref{table:tok_methods} (columns 1-5) demonstrates the phenomenon of joining a word with a combination of prefixes and suffixes, which is extremely productive, that is, can be applied throughout the Hebrew vocabulary.
Consequently, words naturally occur in many different forms, frequently considered as out-of-vocabulary (OOV) by the tokenization, which is known to be a challenge for PLMs \cite{schick2020rare}.

When encountering OOV words, contemporary tokenization methods, which are based on frequencies, tokenize them into sub-words that often lack any morphological meaning.
As a result, PLMs are unable to effectively represent such words based on the their morphological composition \cite{hofmann-etal-2021-superbizarre}.
The significance of handling rare and unseen words through morphological composition is crucial for MRLs, since even with an extensive corpus of text for a given language,\footnote{Hebrew along with other MRLs are considered low to medium-resourced languages.} numerous word forms remain scarce or entirely absent, leaving it to the PLMs to deduce the meaning of such words from its subwords.

Previous studies have identified the tokenization phase as the root cause of this problem \cite{tsarfaty-etal-2019-whats}.
\citet{guetta2022large, feng2022pretraining} further examined the impact of increasing the vocabulary size, demonstrating enhanced performance.
However, this latter method encounters a glass ceiling when facing rare  and OOV words. 
\citet{keren2022breaking, xue-etal-2022-byt5, clark-etal-2022-canine} explore the other extreme by retiring to character-based modeling.
While theoretically holding the potential of learning morphological patterns via the  notion of characters, in practice it achieves on-par results on most morpho-syntactic tasks for Hebrew, and lagging behind in other, more semantic tasks,  like Named-Entity Recognition (NER).
So, while breaking words into characters is a viable option, it is both computationally heavy,  and empirically not quite satisfactory.

\citet{klein-tsarfaty-2020-getting, seker-etal-2022-alephbert} use labeled morphological tokenization at the fine-tuning phase, and show improvements on multiple tasks. However, they do not exploit the potential of employing such tokenization already at the pre-training phase.
Finally, \citet{avraham-goldberg-2017-interplay} display an interplay between morphology and semantics for Hebrew when incorporating morphological knowledge beyond text (e.g. parts-of-speech) at the pre-training phase of the non-contextualized FastText model \cite{bojanowski-etal-2017-enriching}, inspiring us to address the question of employing explicit morphological knowledge into a contextualized PLM for Hebrew.

In this work we explore the impact of incorporating explicit morphological knowledge at the pre-training phase of a contextualized PLM for Hebrew, by introducing morphologically-driven tokenization methods.
Through the utilization of morphologically based sub-word tokenization rather than a purely statistical one, our tokenization methods hold the potential for the model to exploit this morphological knowledge during the acquisition of contextualized representations, as well as to demonstrate morphological composition capabilities at inference time.
By solely modifying the tokenization, no architecture modifications are required, as previous studies have suggested for other MRLs \cite{alkaoud-syed-2020-importance, nzeyimana-niyongabo-rubungo-2022-kinyabert}, making them applicable for other PLMs, e.g., auto-regressive models.

Our results on a benchmark of tasks including: Morphological Parsing, Word-Sense-Disambiguation (WSD), Dependency Parsing, Named entity Recognition (NER) and Question Answering (QA), demonstrate an improvement of morphologically-driven tokenization on most tasks, achieving a notable increase of up-to 3 F1 score for NER and 1 F1 score for dependency parsing.

\section{The Challenge: Bridging the Gap between Morphology \& Tokenization}
\subsection{Hebrew Morphology}
Hebrew is a morphological-rich language, manifesting semantic and syntactic information as modifications of its words form.
For example, the English phrase "and when I loved her" translates into a single Hebrew space-delimited token \<wk/s'hbtyh>, as a result of the following morphological processes: the root \<'hb> (love) is inflected in 1\textsuperscript{st} person-past form, yielding \<'hbty> (\textbf{I} love\textbf{d}); the prefixes combination \<w> (and) \<k/s> (when) are joined, \<wk/s'hbty> (and when I loved); \<yh> (her) suffix is joined, resulting in the complete form (joining the two \<y> into a single one).
Modifications of person, number, and tense, as well as having many exceptional roots, are reflected through non-concatinative morphology \cite{mccarthy1981prosodic}.
Conversely, determining the underlying morphological structure of a word becomes challenging due to the language's high ambiguity arising from the intertwined morphological processes.

\subsection{The Morphological Tokenization Hypothesis}
Our hypothesis is that morphologically based subwords should allow the model to (i) learn more effectively representations of morphologically related words, despite having distinct forms, and (ii) better handle rare and unseen words through morphological composition.
To keep our modifications as minimal as possible, manifested only in changing the tokenization method and not the pre-training architecture, we focus on the morphological phenomenon of joining prefixes and suffixes.

\section{Morphologically-Driven Tokenization}
Putting the aforementioned hypothesis to the test, we consider two different tokenization methods for incorporating explicit morphological knowledge in the pre-training phase: \textit{Morphological Segmentation} and \textit{Prefix-Suffix Separation} tokenizations.

\paragraph{Morphological Segmentation Tokenization}
\label{ref:preseg_seg}
Facing the high ambiguity of Hebrew calls for a process of \textit{morphological segmentation} for tokenizing a word into its morphemes.
In this process, a given word form is disambiguated into its underlying morphemes based on the word's context.
Tokenization of words sharing the same lexical host joined with different prefixes and suffixes results in similar sub-words for this lexical host, as depicted in Table \ref{table:tok_methods} (column 6).
While holding the potential of supplying morphologically based sub-words, the segmentation is done automatically via a dedicated model so this process is prone to segmentation errors propagated to the PLM. Additionally, it introduces dependency of an external segmentation utility used by the PLM in both pre-training and inference phases.

\paragraph{Prefix-Suffix Separation Tokenization}
\label{ref:preseg_presuf}
A lighter alternative for morphologically-based separation of prefixes and suffixes from a word host, is to always separate valid prefixes and suffixes character sequences, as depicted in Table \ref{table:tok_methods} (column 7).
Instead of disambiguating a word form into its underlying morphemes depending on context, we tokenize a word in a deterministic way solely based on its character sequence by separating \textit{potential} prefixes and suffixes.
While this method can still successfully separate prefixes and suffixes from a host, it introduced an additional level of ambiguity, as  words not sharing the same host might be tokenized to the same subwords.

\section{Experiments}
We set out to measure the impact of incorporating explicit morphological knowledge into the pre-training phase through our morphologically driven tokenizations.
We do so by pre-training multiple language models utilizing the different tokenization methods proposed herein (Section \ref{sec:exp_setup}).
Each tokenization method is applied using 3 vocabulary sizes: 16K (small), 32K (standard, \citealp{devlin-etal-2019-bert}) and 64K (large), as a way to assess the impact of the vocabulary size independently of the impact of the tokenization method.
Models are then evaluated on a benchmark of downstream tasks requiring a mixture of morphological, syntactic and semantic knowledge, illuminating the different types of knowledge acquired by models trained on different tokenization methods (Section \ref{sec:benchmark_eval}).

\begin{table*}[t]
\centering
\resizebox{\textwidth}{!}{%
\begin{tabular}{|c|cl|l|c|ccc|cr|cc|}
\hline
 &
  \multicolumn{2}{c|}{NEMO} &
  \multicolumn{1}{c|}{BMC} &
  \begin{tabular}[c]{@{}c@{}}Homographs\end{tabular} &
  \multicolumn{3}{c|}{HTB} &
  \multicolumn{2}{c|}{ParaShoot} &
  \multicolumn{2}{c}{HeQ} \\
 &
  Token &
  \multicolumn{1}{c|}{Morph} &
  \multicolumn{1}{c|}{} &
   &
  POS &
  Features &
  \begin{tabular}[c]{@{}c@{}}Dependency\\ Parsing\end{tabular} &
   &
  \multicolumn{1}{c|}{} &
   &
   \\
 &
  F1 &
  \multicolumn{1}{c|}{F1} &
  \multicolumn{1}{c|}{F1} &
  F1 &
  F1 &
  F1 &
  F1 &
  EM &
  \multicolumn{1}{c|}{F1} &
  EM &
  F1 \\ \hline
Baseline &
  82.99 &
  78.87 &
  90.98 &
  96.04 &
  96.24 &
  95.95 &
  88.34 &
  \multicolumn{1}{r}{15.02} &
  35.86 &
  \textbf{47.31} &
  \textbf{58.55} \\ \cline{1-1}
\begin{tabular}[c]{@{}c@{}}Morphological\\ Segmentation\end{tabular} &
  \textbf{86.43} &
  \textbf{81.54} &
  \textbf{91.09} &
  \textbf{96.10} &
  \textbf{96.39} &
  \textbf{96.10} &
  \textbf{89.31} &
  \multicolumn{1}{r}{\textbf{17.95}} &
  \textbf{38.36} &
  37.72 &
  54.69 \\ \cline{1-1}
\begin{tabular}[c]{@{}c@{}}Prefix-Suffix\\ Separation\end{tabular} &
  85.71 &
  80.71 &
  90.47 &
  95.56 &
  96.26 &
  95.92 &
  87.3 &
  \multicolumn{1}{r}{9.53} &
  24.03 &
  29.22 &
  44.66
 \\ \hline
\end{tabular}%
}
\caption{Main Results: a comparison of all tokenization methods using a vocabulary size of 32K subwords. Best performing method per task is in bold. Full benchmark results can be found in Appendix \ref{appendix:full_results}.}
\label{table:res_main}
\end{table*}

\subsection{Experimental Setup}
\label{sec:exp_setup}
\paragraph{Pre-training}
In order to fairly compare  the different tokenization methods we pre-train BERT-based models using the  dataset of Hebrew Wikipedia and HeDC4 corpus \cite{shalumov2023hero} and the  pre-training configuration with the framework of \cite{izsak-etal-2021-train} (see Appendix \ref{appendix:pre_training_details} for details).

\paragraph{Baseline}
As a baseline, we pre-train models utilizing WordPiece tokenization \cite{schuster2012japanese}, being the standard non-morphologically-driven method.
This follows up on previous successful pre-training of BERT-based models for Hebrew \cite{chriqui2022hebert, seker-etal-2022-alephbert, guetta2022large}.

\paragraph{Morphological Segmentation Tokenization}
We pre-process the dataset using a state-of-the-art morphological segmentation tool \cite{zeldes-2018-characterwise, zeldes-etal-2022-second} in order to convert each word into separated prefixes-host-suffix format.
After this pre-processing takes place, WordPiece tokenization is standardly applied.
Since the Hebrew morphological segmentation phase is applied here in the wild, not on a standard benchmark, we manually compare morphological segmentation tools by evaluating them on a random sample of the pre-training corpus, and selecting the best performing tool (see details in Appendix \ref{appendix:seg_compare}).

\paragraph{Prefix-Suffix Separation Tokenization}
We pre-process the dataset using regular expressions for converting each word into separated prefixes-host-suffix format.
As in the previous method, WordPiece is applied afterwards.

\subsection{Benchmark Evaluation}
\label{sec:benchmark_eval}
We evaluate all models on a Hebrew benchmark that comprises the following tasks, and report their respective standard metrics: Named Entity Recognition (NER) both word- and morpheme-level (F1, \citealp{bareket-tsarfaty-2021-neural, mordecai2005hebrew}); Question Answering (QA) (EM and F1, \citealp{keren-levy-2021-parashoot, cohenheq}); Word Sense Disambiguation (Homographs) (Macro F1, \citealp{shmidman-etal-2023-pretrained});  Morphological segmentation, part-of-speech tagging (POS), morphological features tagging, and dependency parsing (Aligned Multiset F1, UAS F1)\citealp{sade-etal-2018-hebrew, zeldes-etal-2022-second}).
Further details are provided in Appendix \ref{appendix:tasks_description}.

\section{Results \& Analysis}
Our experiments main results are depicted in Table~\ref{table:res_main}. The full results are detailed in Appendix~\ref{appendix:full_results}.

\paragraph{Tokenization Methods Impact}
Both morphologically driven tokenization methods outperform the baseline for NER on NEMO by up-to 3 F1 points in both token and morpheme levels.
The significance of morphologically based tokenization in properly representing OOVs in MRLs is particularly evident in this task as named entities are often unknown to the model during pre-training, and are naturally prefixed with \<w> (and), \</s> (that), \<m200C> (from) and \<l> (to).
Morphological Segmentation tokenization shows an increase of 1 F1 point on dependency parsing, and modest improvements on homographs, POS and morphological features prediction.

The baseline surpasses the Prefix-Suffix Separation method on all tasks except for NER on NEMO, and negligibly for POS on HTB.
While this method proves effective for named entities, it introduces increased ambiguity in regular words, and a higher split count, which has  been previously claimed to decrease models' performance \cite{keren2022breaking, guetta2022large, shmidman-etal-2023-pretrained}.

The weak performance of all models in QA and the contrasting trends on the two datasets makes drawing conclusions regarding the tokenization methods rather challenging.
We leave further investigation of  this issue, which is particular to QA  \cite{keren-levy-2021-parashoot}, for future research.

\begin{figure*}[t]
\centering


\subfigure[NER-token (NEMO)]{
    \includegraphics[width=0.6\textwidth]{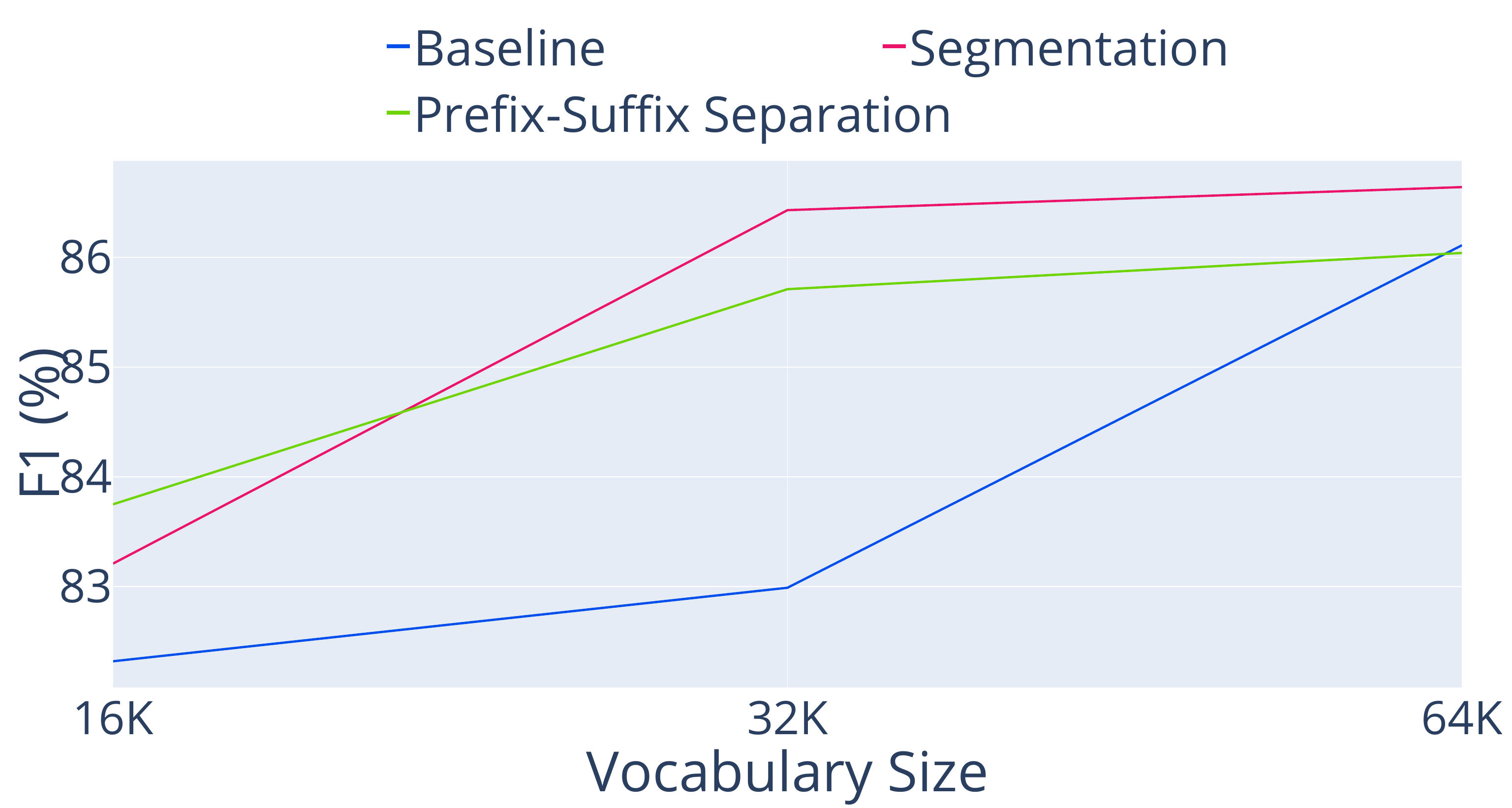}
    \label{fig:res_ner_nemo_token}
}
\hfill
\subfigure[NER-morph (NEMO)]{
    \includegraphics[width=.6\textwidth]{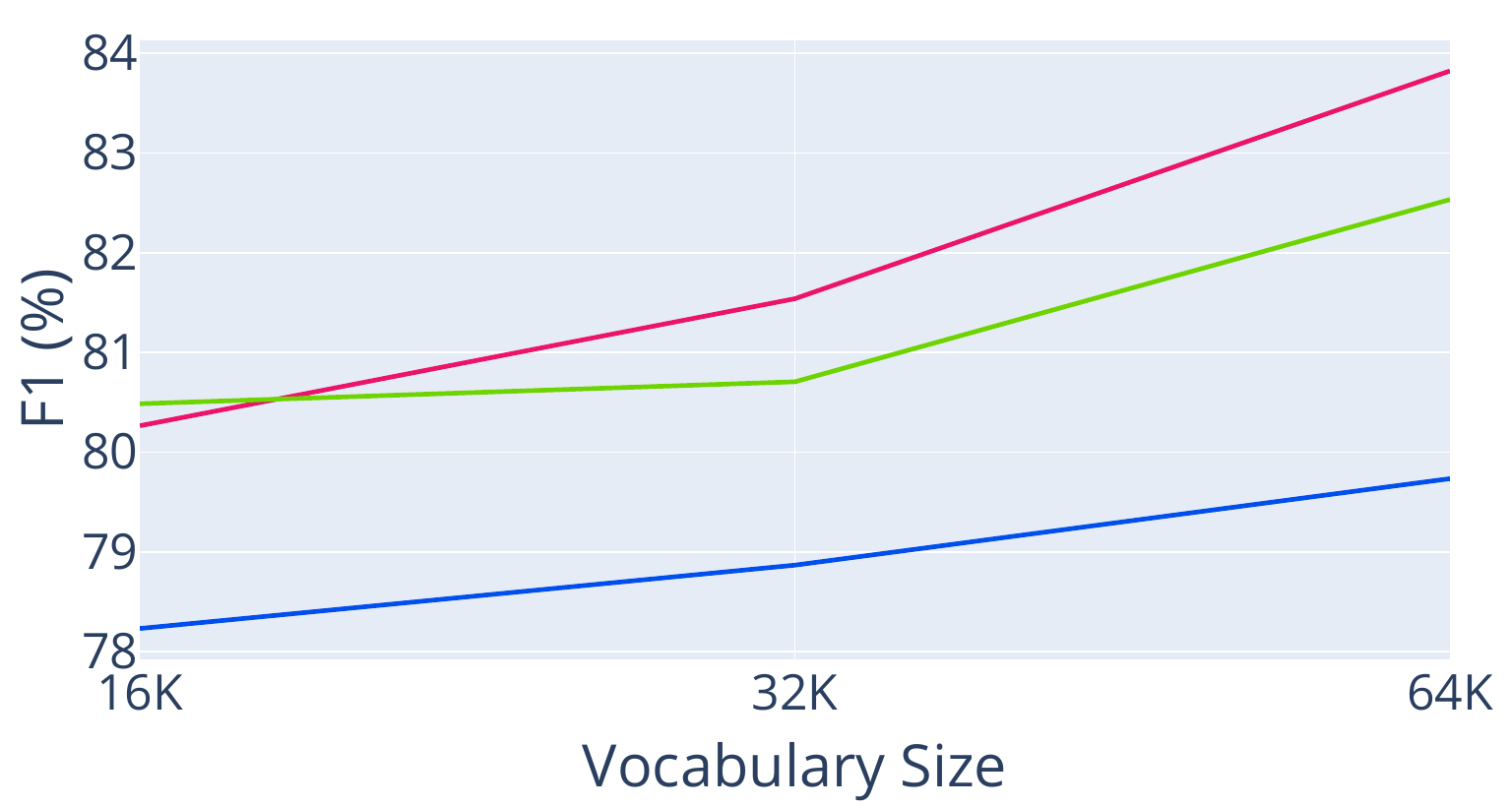}
    \label{fig:res_ner_nemo_morph}
}
\hfill
\subfigure[Dependency Parsing (HTB)]{
    \includegraphics[width=.6\textwidth]{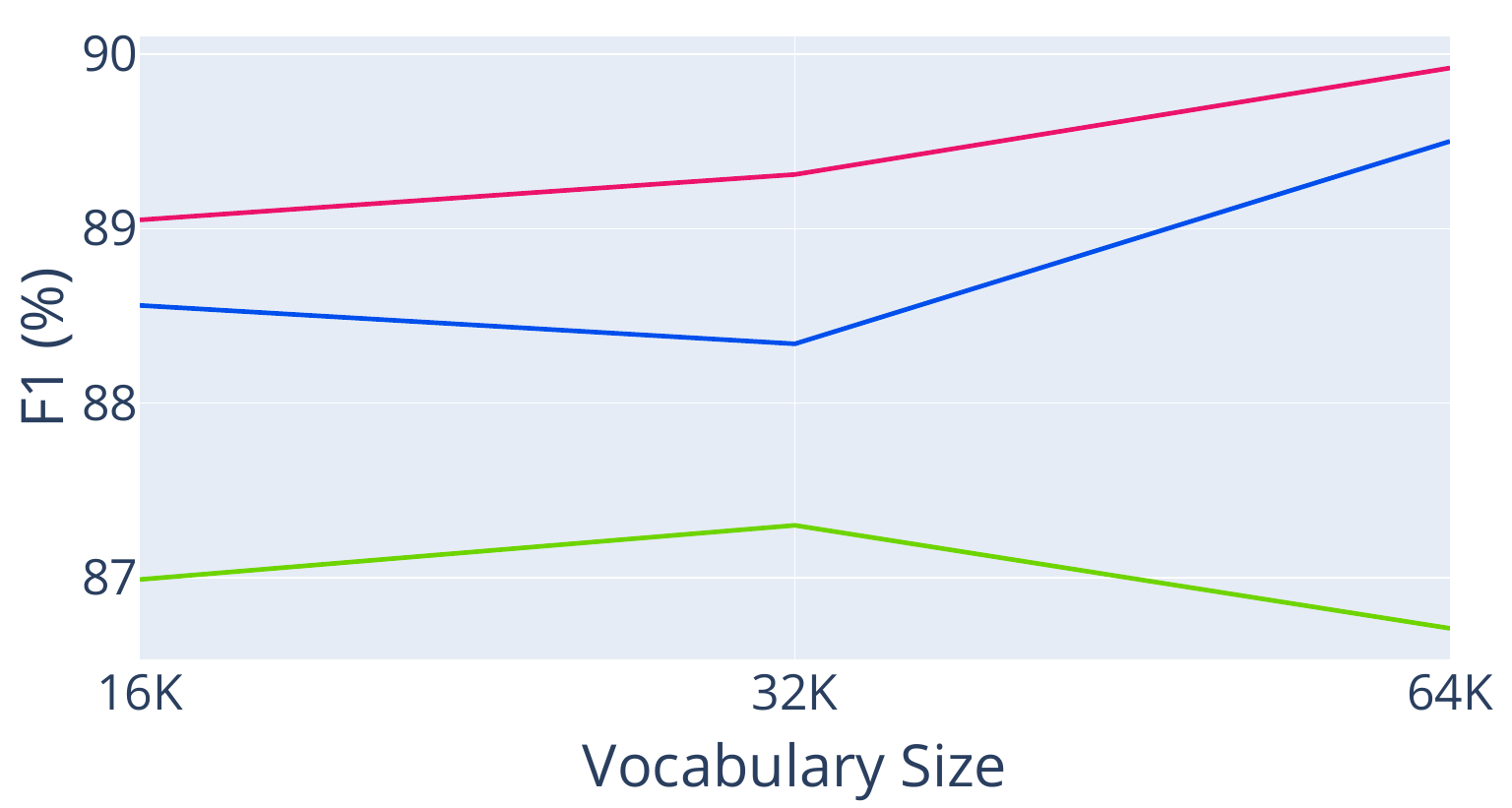}
    \label{fig:res_dep_htb}
}
\subfigure[Dependency Parsing (IAHLTwiki)]{
    \includegraphics[width=.6\textwidth]{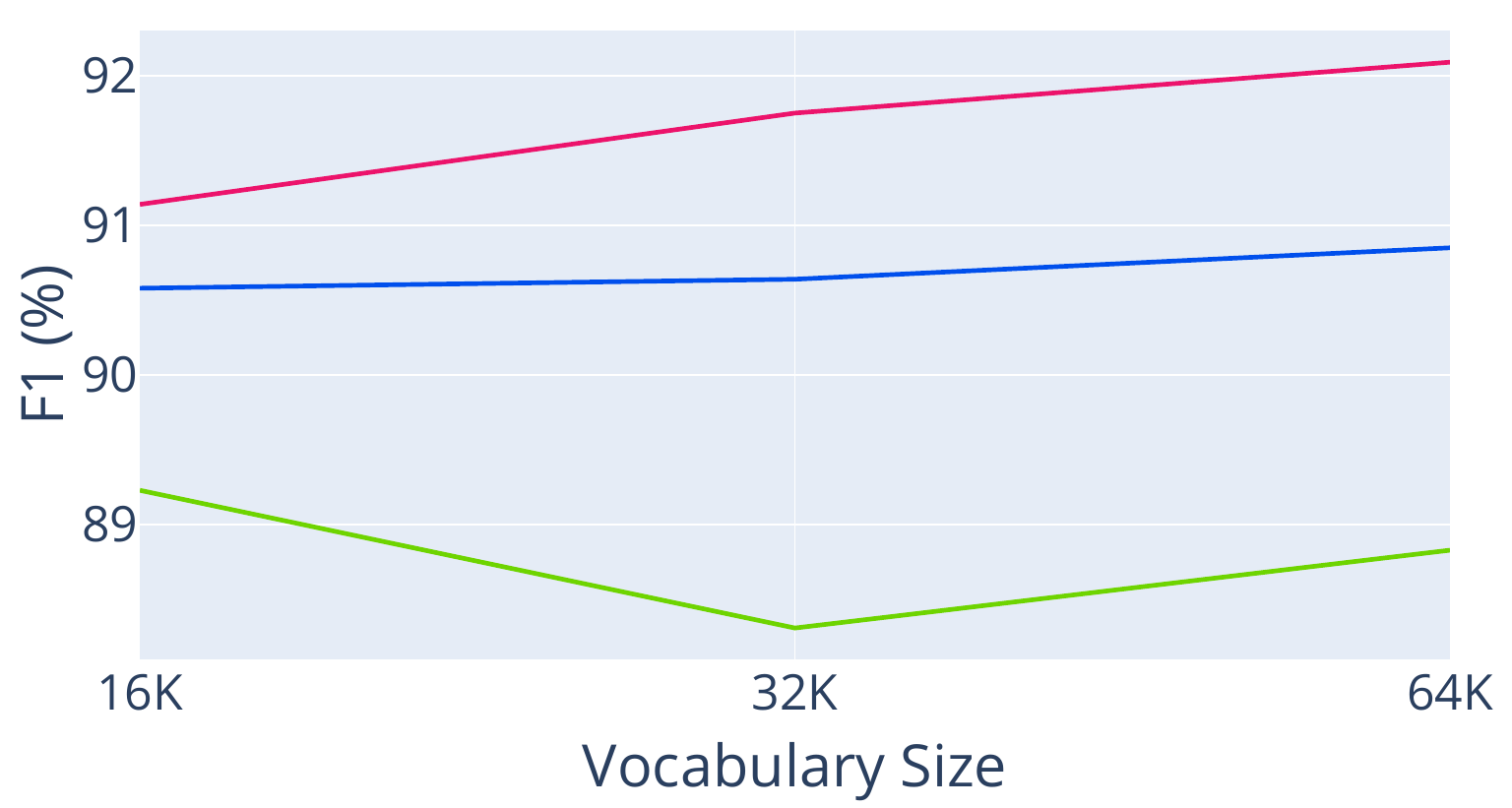}
    \label{fig:res_dep_iahltwiki}
}

\caption{A comparison of the different tokenization methods over the tasks of NER NEMO at both token and morph levels, and dependency parsing on both datasets, across the vocabulary size presented as X axis.}
\label{fig:vocab_size}
\end{figure*}

\paragraph{Vocabulary Size Impact}
Figure \ref{fig:vocab_size} demonstrate the positive impact of increasing the vocabulary size, supporting previous studies \cite{seker-etal-2022-alephbert, guetta2022large}.
This is most evident in NER at token level and QA on both datasets, for almost all tokenization methods (see Appendix \ref{appendix:full_results}).
For NER on NEMO at token level and dependency parsing on HTB, where our Morphological Segmentation tokenization demonstrates the most notable improvement over the baseline with a vocabulary size of 32K, it seems as if increasing the vocabulary size to 64K closes this gap.
We suggest that this is due to memorization rather than generalization to rare and unseen words, inviting future research focusing on the impact of even larger vocabularies \cite{feng2022pretraining}, purely open-vocabulary approaches \cite{tay2021charformer}, as well as on measuring the generalization capabilities of PLMs for Hebrew, as done for other MRLs \cite{moisio-etal-2023-evaluating}.

\section{Conclusion}
Our work illustrates the benefits of incorporating explicit morphological knowledge within the pre-training phase.
Our proposed Morphological Segmentation tokenization method enables the model to effectively learn from and generalize to rare and unseen words through morphological composition.
Experimenting on Hebrew, a highly ambiguous MRL, shows improved performance on multiple tasks, as well as illuminating again the benefits of larger vocabulary sizes.
These findings call for research endeavours focusing on better tokenization methods for better language models for MRLs.

\section*{Limitations}
Our usage of \citet{izsak-etal-2021-train} as a pre-training framework explores a single default setup with respect to the architecture (BERT-based), model's size (large), and other pre-training parameters.
While being the first apples-to-apples comparison of Hebrew language models, as well as an effective and fast pre-training utility, it might still be insufficient for achieving optimal  model performance.

Another limitation of our proposed Morphological Segmentation tokenization is its inherent reliance on a pre-processing phase which is prone to errors, potentially having a negative impact on the learned representations, as well as in inference time.
Additionally, while introducing a more morphological adequate tokenizations, they yield higher number of subwords per word, which might undermine the models' performance as previously argued by \citet{guetta2022large, shmidman-etal-2023-pretrained}.

From a morphological perspective, our research is focused on incorporating morphological knowledge of prefix and suffix nature only, whereas Hebrew morphologically shows far richer phenomena (Appendix \ref{appendix:heb_morph_beyond}), which we leave for future research.

\section*{Acknowledgements}
We thank our anonymous reviewers for their helpful comments on this paper.
This research was funded by the Israeli Ministry of Science and Technology (MOST) grant No.\ 3-17992, and an Israeli Innovation Authority grant (IIA) KAMIN grant. In addition, This project has received funding from the European Research Council (ERC) under the European Union's Horizon 2020 research and innovation programme, grant agreement No.\ 677352.
The compute for this research was provided by the Data Science Institute (DSI) at Bar Ilan University and a VATAT grant for which we are grateful.

\bibliography{anthology,custom}
\bibliographystyle{acl_natbib}

\appendix

\section{Hebrew Prefixes-Host-Suffix Details}
\paragraph{Hebrew Prefixes and Suffixes}
Hebrew prefixes posses the role of functional words: \<m/s> (since), \<k/s> (when), \<b> (in), \<l> (to), \<k200C> (as/like), \<w> (and), \<h> (the), \</s> (that), \<m200C> (from).
Combining multiple prefixes (e.g. \<k/s>+\<w> (and when), \<h>+\<m200C>+\</s> (that from the) results in beyond 55 different valid forms.

Hebrew suffixes indicate either genitive or accusative case-marking for nouns and verbs, respectively: \<y> (mine), \<K> (you/rs 2nd person), \<h> (her/s), \<w> (him/his), \<nw> (ours/us), \<kN> (you/rs 3rd person, feminine), \<kM> (you/rs 3rd person, masculine), \<N>/\<hN> (them/theirs, feminine), \<M>/\<hM> (them/theirs, masculine). Following its linguistic functionality, only a single suffix can be joined with a word.

\label{appendix:heb_morph_beyond}
\paragraph{Hebrew's Morphology Beyond Prefixes-Host-Suffix}
Hebrew includes also the processes of derivation and inflection, which are non-concatinative.
Taking them into account, tokenization could be potentially further improved by working on a word's root-granularity rather than on a host granularity separated from its prefixes and suffix only, allowing to further generalize to nouns' singular-plural forms and verb's person, tense and \textit{Binyan} (Hebrew has 7 different verb structures called Binyan, translated as structure or building in Hebrew, reflecting linguistic information as passive-active, voice and more.). However, existing Hebrew tools and datasets do not provide this granularity.
Hebrew also has a complete system of diacritics, efficiently disambiguating most texts.
However, the majority of natural and available text is non-diacritized, thus it can not be exploited.

\paragraph{WordPiece Integration}
Since WordPiece tokenization ignores spaces \footnote{We refer to Huggingface's implementation of a pre-tokenizer, pre-processing text into tokens and ignoring spaces, applying WordPiece on the tokens without considering the spaces.} and due to the fact Hebrew prefixes like \<w> (and) and \<h> (the) are also used as suffix (his and hers, respectively), separation of prefixes and suffixes might result in a sentence-level ambiguity, where a word's suffix might account as the next word's prefix.
To avoid such ambiguity we mark prefixes as \textit{p+} distinguished from suffixes which we mark as \textit{+s}.

\paragraph{Hebrew Overlapping Prefixes and Suffixes}
Hebrew prefixes \<m/s> (since) and \<m200C> (from), \<k/s> (when) and \<k200C> (as/like), and suffix \<nw> (our) and \<w> (his), have a common prefix/suffix, respectively.
Due to Hebrew's extreme ambiguity, a word like \<m/srwt> beginning with the prefix \<m/s> might be a prefix \<m/s> joined to a host (since Ruth), or a prefix \<m200C> joined to a host beginning with \</s> (as a service), or simply a host beginning with \<m/s> (jobs).
Therefore we further break \<m/s> as \</s> \<m200C>, \<k/s> as \</s> \<k200C>, and \<nw> as \<w> \<n200C>, to achieve the maximal overlap of subwords between words.

\paragraph{Splitting Only Valid Prefixes}
As depicted in Table \ref{table:tok_methods} (column 7), the Prefix-Suffix Separation Tokenization does not reach maximal overlap because the word \</s/s.hrwr> is tokenized into \</s.hrwr>+\</s> instead of into \<.hrwr>+\</s>+\</s> which includes the host's subword \<.hrwr> as in the other forms.
Since Hebrew morphology permits a mixture of up to 4 different prefixes (e.g. \<k200C>+\<b>+\</s>+\<w>
 like in the word [...\<pwrsM ky>] \<w/sbkm.h.syt> [...\<mhmqryM>] translated as [It has been published that...] and that in about half [of the incidents...]), we chose to separate only 55 valid combination of prefixes, instead of all possible prefixes, due to the exponentially large combinatorical space of possible mixture of prefixes (beyond 3K), making it closer to a character-based approach, possibly obscuring the actual impact of incorporating morphological knowledge, rather than utilize characters.

\section{Morphological Segmentation Tools Comparison}
\label{appendix:seg_compare}
We consider 4 different Hebrew morphological segmentation tools: YAP \cite{seker-etal-2018-universal} which is based on a lexicon, RFTokenizer \cite{zeldes-2018-characterwise} which is character based, optionally considering PLM representations, and assuming concatinative morphology, and the PLM based, not assuming concatinative morphology, Trankit \cite{nguyen-etal-2021-trankit}, and \cite{brusilovsky2022neural}.

We do not use \cite{nguyen-etal-2021-trankit} since its segmentation is limited to either keeping a word as is, or segmenting it in a single way, disregarding Hebrew's far higher ambiguity including many words with more than 2 possible segmentations \cite{shmidman-etal-2023-pretrained}.
We choose to not use \cite{brusilovsky2022neural} despite its state-of-the-art performance because of its hallucinations, most notably for names, which are abundant in natural free texts used for pre-training.

We segment 128 sentences (3K words) randomly sampled from the pre-training corpus using both YAP and RFTokenizer.
Automatic inspection reveals they produce identical segmentations for 92\% of the words, excluding non-words and truly ambiguous cases.
When they disagree on the segmentation, manual inspection reveals RFTokenizer presents an error rate of 4\% whereas YAP demonstrates an error rate of 52\%.

Beyond the performance aspect, another important consideration is the computation time required to process the whole dataset using each of these tools.
RFTokenizer computation is fast enough to allow pre-processing the whole dataset in reasonable time by parallelizing the pre-processing over multiple GPUs.

\section{Pre-training Details}
\label{appendix:pre_training_details}
\paragraph{Dataset} Pre-training dataset include Hebrew Wikipedia (1.4GB) and the recently released HeDC4 corpus \cite{shalumov2023hero} containing 47.5GB of de-duplicated cleaned texts.
The dataset is pre-masked using 5 copies, yielding a little more than 100GB of text, as recommended.
1\% of the dataset is held for evaluation along the pre-training.

\paragraph{Pre-training Parameters} Pre-training follows 
\cite{izsak-etal-2021-train} recommendations, using 23K steps of 4K batch size, instead of the time-based budget, roughly achieving pre-training of 96M samples in total.
Rest of the parameters are as indicated in the paper and in their github repository: \url{https://github.com/IntelLabs/academic-budget-bert}.

\section{Downstream Tasks Details}
\label{appendix:tasks_description}

\subsection{Fine-tuning Details}
For NER at token level and QA we fine-tune all models using Huggingface's framework \cite{wolf-etal-2020-transformers} token classification and question answering standard implementations.
For morphological-level tasks of segmentation, part-of-speech, morphological features prediction and NER we fine-tune all models \cite{brusilovsky2022neural} framework for jointly learning segmentation \& POS \& morphological features prediction and segmentation \& NER.
Since Morphological Segmentation Tokenization requires pre-segmentation, which is performed by a model fine-tuned by itself on a morpheme-level dataset, we use separate RFTokenizer models for the different datasets: for UD-HTB \cite{sade-etal-2018-hebrew} we use an RFTokenizer fine-tuned on UD-IAHLT \cite{zeldes-etal-2022-second} only, and vice-versa.
For dependency parsing we use DiaParser \cite{attardi-etal-2021-biaffine} following \cite{zeldes-etal-2022-second} evaluation, provided gold segmentation.
For Homographs we produce PLMs representations of the homographs in context (i.e. effectively equivalent to full training with frozen PLMs), and use \cite{pedregosa2011scikit} to train a separate MLP on top of the sum of the representations (as each homograph might be tokenized into more than one subword) for each homograph, using either 5, 25, 100, or 90\% of the dataset for training, and testing on the rest, using a 10-fold cross validation, as recommended by \cite{shmidman-etal-2023-pretrained}.

\subsection{Pre-processing Using Proposed Tokenization}
From implementation perspective, our tokenization methods are applied on the dataset, rather than incorporated into the PLM's Tokenizer \cite{wolf-etal-2020-transformers}.
For NER and the morphologic tasks, this takes the form of applying the tokenizations of RFTokenizer and our regex based tokenization on the sentences tokens.
For Homographs, we run the tokenizers on the datasets sentences.
QA requires further adaptation as the context, question and answer change.
First, we run our tokenization on the context and on the question separately.
Then, we construct a regex on top of the original answer, consuming potential separators between prefix to host and between host to suffix, added by our tokenizations.
We use this regex to search the newly tokenized context, to find the new form of the answer and its position, following the dataset's format.
This ensures we do not tokenize the context and the answer in different ways, which is possible since RFTokenizer utilizes the surrounding words of the answer, which is different in the context and in the extracted labeled answer.

\section{Benchmark Full Results}
\label{appendix:full_results}
We hereby present the full results of all models, including all tasks, all tokenization methods and all vocabulary sizes.
Each table refers to a different task, except Table \ref{table:res_morph} which includes all morphologic tasks together.



\begin{table*}[]
\centering
\resizebox{\textwidth}{!}{
\begin{tabular}{cc|cll}
                                                                                      &     & NEMO-token & \multicolumn{1}{c}{NEMO-morph} & \multicolumn{1}{c}{BMC-token} \\
\begin{tabular}[c]{@{}c@{}}Tokenization\\ Method\end{tabular} &
  \begin{tabular}[c]{@{}c@{}}Vocabulary\\ Size\end{tabular} &
  F1 &
  \multicolumn{1}{c}{F1} &
  \multicolumn{1}{c}{F1} \\ \hline
\multirow{3}{*}{Baseline}                                                             & 16K & 82.32      & 78.23                          & 90.06                         \\
                                                                                      & 32K & 82.99      & 78.87                          & 90.98                         \\
                                                                                      & 64K & 86.11      & 79.73                          & 91.64                         \\ \hline
\multirow{3}{*}{\begin{tabular}[c]{@{}c@{}}Morphological\\ Segmentation\end{tabular}} & 16K & 83.21      & 80.26                          & 90.42                         \\
                                                                                      & 32K & 86.43      & 81.54                          & 91.09                         \\
                                                                                      & 64K & 86.64      & 83.82                          & 91.31                         \\ \hline
\multirow{3}{*}{\begin{tabular}[c]{@{}c@{}}Prefix-Suffix\\ Separation\end{tabular}}   & 16K & 83.75      & 80.49                          & 89.47                         \\
                                                                                      & 32K & 85.71      & 80.71                          & 90.47                         \\
                                                                                      & 64K & 86.04      & 82.53                          & 90.47                        
\end{tabular}}
\caption{NER Results}
\label{table:res_ner}
\end{table*}

\begin{table*}[]
\centering
\resizebox{\textwidth}{!}{%
\begin{tabular}{cc|ccccc|ccccc}
 &
   &
  \multicolumn{5}{c|}{UD-HTB \& NEMO} &
  \multicolumn{5}{c}{UD-IAHLTwiki} \\
 &
   &
  Seg &
  POS &
  Features &
  \multicolumn{2}{c|}{\begin{tabular}[c]{@{}c@{}}Dependency\\ Parsing\end{tabular}} &
  Seg &
  POS &
  Features &
  \multicolumn{2}{c}{\begin{tabular}[c]{@{}c@{}}Dependency\\ Parsing\end{tabular}} \\
\begin{tabular}[c]{@{}c@{}}Tokenization\\ Strategy\end{tabular} &
  \begin{tabular}[c]{@{}c@{}}Vocabulary\\ Size\end{tabular} &
  F1 &
  F1 &
  F1 &
  UAS &
  LAS &
  F1 &
  F1 &
  F1 &
  UAS &
  LAS \\ \hline
\multirow{3}{*}{Baseline} &
  16K &
  98.24 &
  96.28 &
  95.86 &
  91.79 &
  88.56 &
  98.09 &
  95.70 &
  92.77 &
  93.75 &
  90.58 \\
 &
  32K &
  98.16 &
  96.24 &
  95.95 &
  91.64 &
  88.34 &
  98.16 &
  95.92 &
  93.03 &
  93.91 &
  90.64 \\
 &
  64K &
  98.05 &
  96.08 &
  95.84 &
  92.43 &
  89.5 &
  98.07 &
  95.80 &
  92.83 &
  93.96 &
  90.85 \\ \hline
\multirow{3}{*}{\begin{tabular}[c]{@{}c@{}}Morphological\\ Segmentation\end{tabular}} &
  16K &
  98.20 &
  96.40 &
  96.04 &
  92.29 &
  89.05 &
  98.00 &
  95.71 &
  92.80 &
  94.29 &
  91.14 \\
 &
  32K &
  98.22 &
  96.39 &
  96.10 &
  92.43 &
  89.31 &
  98.09 &
  95.99 &
  93.14 &
  94.63 &
  91.75 \\
 &
  64K &
  98.16 &
  96.21 &
  95.97 &
  92.98 &
  89.92 &
  97.64 &
  95.50 &
  92.73 &
  94.81 &
  92.09 \\ \hline
\multirow{3}{*}{\begin{tabular}[c]{@{}c@{}}Prefix-Suffix\\ Separation\end{tabular}} &
  16K &
  98.20 &
  96.37 &
  96.05 &
  90.58 &
  86.99 &
  97.79 &
  95.42 &
  92.57 &
  92.73 &
  89.23 \\
 &
  32K &
  98.28 &
  96.26 &
  95.92 &
  90.82 &
  87.3 &
  97.49 &
  95.25 &
  92.48 &
  91.73 &
  88.31 \\
 &
  64K &
  98.17 &
  96.33 &
  96.03 &
  90.39 &
  86.71 &
  97.89 &
  95.73 &
  92.83 &
  92.44 &
  88.83
\end{tabular}%
}
\caption{Morphologic tasks results}
\label{table:res_morph}
\end{table*}

\begin{table*}[]
\centering
\resizebox{\textwidth}{!}{
\begin{tabular}{cc|cc}
                                                                                      &     & ParaShoot     & HeQ           \\
\begin{tabular}[c]{@{}c@{}}Tokenization\\ Method\end{tabular} & \begin{tabular}[c]{@{}c@{}}Vocabulary\\ Size\end{tabular} & F1 / EM & F1 / EM \\ \hline
\multirow{3}{*}{Baseline}                                                             & 16K & 26.71 / 10.73 & 50.13 / 38.65 \\
                                                                                      & 32K & 35.86 / 15.02 & 58.55 / 47.31 \\
                                                                                      & 64K & 40.87 / 19.54 & 62.94 / 52.22 \\ \hline
\multirow{3}{*}{\begin{tabular}[c]{@{}c@{}}Morphological\\ Segmentation\end{tabular}} & 16K & 24.27 / 08.68 & 46.53 / 30.19 \\
                                                                                      & 32K & 38.36 / 17.95 & 54.69 / 37.72 \\
                                                                                      & 64K & 29.75 / 11.41 & 56.78 / 39.92 \\ \hline
\multirow{3}{*}{\begin{tabular}[c]{@{}c@{}}Prefix-Suffix\\ Separation\end{tabular}}   & 16K & 27.67 / 14.26 & 41.95 / 27.48 \\
                                                                                      & 32K & 24.03 / 09.53 & 44.66 / 29.22 \\
                                                                                      & 64K & 30.13 / 13.01 & 46.44 / 29.61
\end{tabular}}
\caption{QA results}
\label{table:res_qa}
\end{table*}

\begin{table*}[]
\centering
\resizebox{\textwidth}{!}{%
\begin{tabular}{cc|c|ccc}
\begin{tabular}[c]{@{}c@{}}Tokenization\\ Method\end{tabular} &
  \begin{tabular}[c]{@{}c@{}}Vocabulary\\ Size\end{tabular} &
  \begin{tabular}[c]{@{}c@{}}F1\\ k = 90\%\end{tabular} &
  \begin{tabular}[c]{@{}c@{}}F1\\ k = 5\end{tabular} &
  \begin{tabular}[c]{@{}c@{}}F1\\ k = 25\end{tabular} &
  \begin{tabular}[c]{@{}c@{}}F1\\ k = 100\end{tabular} \\ \hline
\multirow{3}{*}{Baseline}                                                             & 16K & 95.57 & 63.48 & 86.49 & 92.23 \\
                                                                                      & 32K & 96.04 & 66.62 & 88.84 & 93.63 \\
                                                                                      & 64K & 95.64 & 64.88 & 88.15 & 93.15 \\ \hline
\multirow{3}{*}{\begin{tabular}[c]{@{}c@{}}Morphological\\ Segmentation\end{tabular}} & 16K & 96.09 & 68.91 & 90.16 & 94.05 \\
                                                                                      & 32K & 96.10 & 63.95 & 87.03 & 93.14 \\
                                                                                      & 64K & 96.20 & 68.41 & 90.11 & 94.17 \\ \hline
\multirow{3}{*}{\begin{tabular}[c]{@{}c@{}}Prefix-Suffix\\ Separation\end{tabular}}   & 16K & 95.20 & 64.23 & 86.81 & 92.16 \\
                                                                                      & 32K & 95.56 & 66.73 & 88.79 & 93.11 \\
                                                                                      & 64K & 95.34 & 64.92 & 87.85 & 92.72
\end{tabular}%
}
\caption{Homographs results}
\label{table:res_homographs}
\end{table*}

\end{document}